\documentclass[preprint,12pt]{elsarticle}

\usepackage{amssymb}
\usepackage{amsmath}
\usepackage{graphicx}
\usepackage{epstopdf}
\usepackage{subfigure}
\usepackage{siunitx}
\usepackage{multirow}
\usepackage{booktabs}
\usepackage{hyperref}
\usepackage{mathrsfs}
\journal{arXiv}

\begin{document}

\begin{frontmatter}



\title{Option Pricing Using Ensemble Learning} 


\author{Zeyuan Li\textsuperscript{a},   Qingdao Huang\textsuperscript{a}} 

\affiliation{organization={School of Mathematics, Jilin University},
            addressline={2699 Qianjin Street, Changchun, Jilin Province}, 
            city={Changchun},
            postcode={130012},
            state={Jilin Province},
            country={China}}
\begin{abstract}
Ensemble learning is characterized by flexibility, high precision, and refined structure.
As a critical component within computational finance, option pricing with machine learning requires both high predictive accuracy and reduced structural 
complexity—features that align well with the inherent advantages of ensemble learning.
This paper investigates the application of ensemble learning to option pricing, and conducts a comparative analysis with classical machine learning models 
to assess their performance in terms of accuracy, local feature extraction, and robustness to noise.
A novel experimental strategy is introduced, leveraging parameter transfer across experiments to improve robustness and realism in financial simulations.
Building upon this strategy, an evaluation mechanism is developed that incorporates a scoring strategy and a weighted evaluation strategy 
explicitly emphasizing the foundational role of 
financial theory. This mechanism embodies an orderly integration of theoretical finance and computational methods.
In addition, the study examines the interaction between sliding window technique and noise, revealing nuanced patterns that suggest a potential connection 
relevant to ongoing research in machine learning and data science.
\end{abstract}


\end{frontmatter}




\section{Introduction}
Option pricing using machine learning is an emerging interdisciplinary field that leverages data-driven models to enhance traditional financial theory. 
Its core objectives—achieving both precision and generalization—are closely aligned with fundamental goals in computer science, 
while also retaining distinct characteristics rooted in financial theory.

Since the late 20th century, decision tree algorithms have been proposed as effective tools for classification and regression tasks\cite{breiman2017classification}\cite{guo1992classification}. 
These methods have evolved into ensemble learning techniques, which are capable of addressing increasingly complex problems due to their inherent flexibility\cite{breiman2001random}\cite{polikar2012ensemble}.

Ensemble learning is particularly characterized by its flexibility, high precision, and structural sophistication. Compared to traditional neural networks, 
it typically requires less training time and is easier to implement, 
making it especially well-suited for option pricing applications within the machine learning framework.

Scholars in the field of option pricing with machine learning have primarily focused on neural networks, especially since J.M. Hutchinson et al. 
and M. Malliaris et al. introduced the core concept of nonparametric methods using neural networks 
in the late 1990s\cite{hutchinson1994nonparametric}\cite{malliaris1993neural}. 
This approach was further refined by U. Anders et al. in 1998\cite{anders1998improving} and subsequently extended by numerous researchers\cite{taudes1998real}\cite{yao2000option}\cite{bennell2004black}.

Based on recent research trends and the potential scope of future studies, three prominent directions have emerged in this field:
(1) applying state-of-the-art neural network frameworks for performance improvement\cite{ge20213d}\cite{li2024option}; 
(2) expanding the application of neural networks to broader machine learning paradigms\cite{ivașcu2021option}; 
and (3) re-examining the financial properties of option pricing to develop more appropriate strategies grounded in domain-specific theory.

Building on the second and third directions mentioned earlier,
this study deepens the integration of machine learning and financial thinking in option pricing.
Inspired by Ivașcu et al.\cite{ivașcu2021option}, ensemble learning models are used as the core framework, supported by other machine learning methods.
This addresses a common research gap where ensemble methods are often overlooked, offering a new perspective for future work.
By leveraging their structural advantages and flexibility, ensemble models enhance performance and reduce over-regularization.

Beyond model choice, this research proposes and applies a novel evaluation mechanism grounded in classical financial theory.
Two major experiments are designed to demonstrate the Black–Scholes model’s role as a theoretical complement in machine learning-based option pricing.
This highlights that financial models are not merely accuracy benchmarks but essential tools providing theoretical context.

The main contributions of this study are as follows:

A novel experimental strategy is constructed to enhance the robustness of results. 
By incorporating the requirement for generalization capability, the experiments are brought closer to real-world financial environments.

Sliding window and noise contrast experiments are conducted to explore the potential relationship between models’ ability to extract local information and their robustness to noise.

This study is the first to propose a scoring strategy and a weighted evaluation strategy for option pricing using machine learning. 
This evaluation mechanism enable financial theory to guide and participate in the pricing process of nonparametric models.

This research is the first to adopt ensemble learning as the central modeling framework in this field. 
A series of experiments is conducted to evaluate the models, with results showing that ensemble models such as NGBoost, LGBM, and others express strong performance.

The paper is structured as follows: Chapter 2 provides a brief overview of ensemble learning models. 
Chapter 3 presents both the novel and existing strategies utilized in this study. 
Chapters 4 and 5 cover the experimental design, results, and discussion. Finally, Chapter 6 concludes the paper with a summary and future research directions.
\section{Models description}
In this section, ensemble learning models are mainly introduced, and other models are briefly introduced. 
\subsection{Random Forest}
Random Forest(RF) is widely recognized as a representative ensemble learning technique. It was first introduced by Breiman et al. in 2001\cite{breiman2001random}. 
Based on the Bagging(Bootstrap Aggregating) framework, RF employs decision trees as base learners and incorporates random feature selection at each node, 
which is a defining characteristic of the method. RF exhibits two forms of perturbation: sample perturbation, introduced by Bagging, and attribute perturbation, 
introduced by random feature selection.

Sample perturbation refers to the process of drawing random bootstrap samples from the original training set, which is the core idea of Bagging. 
Attribute perturbation, on the other hand, involves randomly selecting a subset of $k$ features(given a total of $d$ features, where $d \geq k$) at each node 
and then allowing the decision tree to select the optimal feature from this subset to perform the split.

These perturbation properties rely on the increased diversity among individual learners to enhance the generalization capability of the overall model. 
RF is known for its simple structure, computational efficiency, and ease of implementation.

\subsection{XGBoost}
Boosting is a widely used statistical learning method that embodies the concept of ensemble learning. It typically employs decision trees as base learners, 
commonly referred to as boosting trees.
To address this problem, Friedman proposed the gradient boosting method, which uses the steepest descent approach for function approximation \cite{friedman2001greedy}.
This strategy leverages the negative gradient of the loss function to approximate the residuals, as shown below:
\begin{equation}
-[\frac{\partial L(y_{i},f(x_{i}))}{\partial f(x_{i})} ]_{f(x)=f_{m-1}(x)} 
\end{equation}
where $L(\cdot)$ denotes the loss function. The training dataset is defined as \\
$T = \{(x_{1}, y_{1}), (x_{2}, y_{2}), \cdots, (x_{N}, y_{N})\}$,
where $y_{i} \in \mathcal{Y} \subseteq \mathbf{R}$, and $x_{i} \in \mathcal{X} \subseteq \mathbf{R}^n$.

However, this strategy consider the objective function $L(y,f(x_{i}))$ without regularization term, thus cannot avoid risk of regularization. XGBoost
(eXtreme Gradient Boosting) is a method consider regularization term since its objective function is as shown\cite{chen2016xgboost}:
\begin{equation}
  \mathscr{L}(\phi ) = \sum_{i}l(y_{i},\hat{y_{i}}) + \sum_{k}\Omega (f_{k})
\end{equation}  
where $\sum_{k}\Omega (f_{k})=\sum_{k=1}^{K}\Omega (f_{k})=\Omega (f_{t})+constant$ since in time $t$.
By taking Taylor expansion of the objective and with constants removed, objective function has\cite{chen2014introduction}:
\begin{equation}
  \mathscr{L}^{(t)} = \sum_{i=1}^{n}[g_{i}f_{t}(x_{i})+\frac{1}{2}h_{i}f_{t}^{2}(x_{i})] + \sum_{k}\Omega (f_{k})
\end{equation} 
where $g_{i}=\partial_{\hat{y}^{(t+1)}}l(y_{i},\hat{y}^{(t+1)}), h_{i}=\partial_{\hat{y}^{(t+1)}}^{2}l(y_{i},\hat{y}^{(t+1)})$.

XGBoost is an effective ensemble learning model, 
as it not only applies Taylor expansion to improve accuracy, but also incorporates regularization to prevent overfitting and simplify the model structure.
\subsection{LGBM}
Another boosting method, called LGBM, also known as LightGBM(Light Gradient Boosting Machine)\cite{ke2017lightgbm}, employs the negative gradient of the loss function to simulate residuals.
LGBM is known for its speed, efficiency, low memory usage, and high accuracy. 
Both LGBM and XGBoost are widely used and well-regarded by experts in the field of machine learning\cite{massaoudi2021novel}\cite{sajid2023ensemble}.

The key differences between XGBoost and LGBM include, among others, 
that LGBM retains large samples while randomly selecting a subset of small samples, and further enhances the information gain from these small samples. 

LGBM incorporates several optimizations \cite{ke2017lightgbm}, including: 
a histogram-based algorithm for feature binning; 
leaf-wise with max depth limitation; 
enhanced cache hit rate; support for categorical features; 
acceleration based on histogram differences; 
optimization for sparse features using histogram representation; and multithreading to enable parallel computation.

\subsection{DeepForest}
DeepForest, also known as gcForest(multi-Grained Cascade Forest), has been proposed as a promising alternative to deep neural networks\cite{zhou2019deep}.
Compared to traditional deep learning methods, it offers easier training and demonstrates strong performance, particularly in experiments involving small sample sizes.
However, despite being an ensemble-based approach, 
DeepForest introduces additional structural complexity compared to conventional ensemble models due to its multi-layer cascade forest architecture.

The cascade forest structure is a novel architecture proposed in the study, consisting of two types of forests: a traditional random forest(RF) and a completely-random tree forest.
In summary, DeepForest extends the construction of completely-random trees introduced in previous work\cite{mu2017classification}\cite{liu2008spectrum}, 
and aims to approximate the output behavior of neural networks.

Another distinctive feature of this model is the multi-grained scanning mechanism, which is inspired by the Bagging. 
It utilizes a sliding window to generate additional training instances, which are then fed into both types of forests for representation learning.
\subsection{CatBoost}
CatBoost (Categorical Boosting) is one of the mainstream models based on the GBDT(Gradient Boosting Decision Tree) framework. 
Compared to XGBoost and LGBM, CatBoost employs oblivious decision trees as base learners, 
which involve fewer parameters and provide better generalization performance. 
The name “CatBoost” reflects its core design: a combination of categorical feature handling and boosting, 
specifically optimized for efficient and accurate processing of categorical variables\cite{dorogush2018catboost}.

CatBoost incorporates several notable features. First, it provides native handling of categorical features, 
allowing it to process categorical variables directly without complex preprocessing. 
This is achieved through a specialized encoding strategy that integrates target statistics while applying smoothing techniques to reduce the risk of overfitting.

Second, it employs ordered boosting, where trees are constructed sequentially in a way that preserves the information of GBDT. 
Each tree only utilizes the prediction results from its predecessors.

Third, the oblivious tree structure used in CatBoost ensures that all splits at each level of a tree are made using the same feature and threshold. 
This architectural choice enhances robustness, and enables faster prediction speed.

Finally, CatBoost adopts a dynamic learning rate mechanism, where the learning rate is adaptively adjusted based on the number of boosting iterations.
\subsection{NGBoost}
Inspired by Martens' work on the natural gradient method \cite{martens2014new}, Duan et al. developed a formal mathematical framework—specifically, 
a generalized natural gradient—to propose NGBoost (Natural Gradient Boosting) \cite{duan2020ngboost}.

NGBoost leverages proper scoring rules—for example, Maximum Likelihood Estimation (MLE) and Continuous Ranked Probability Score (CRPS)
—in combination with the generalized natural gradient, 
which makes it practical to estimate the relevant parameters $\theta$ in order to generate probabilistic predictions $P_{\theta }(y|x)$, 
thereby enabling the model to return every predictive probability distribution.

The generalized natural gradient is particularly advantageous because it represents the steepest ascent direction in Riemannian space. 
By employing the natural gradient for parameter learning, NGBoost makes the optimization problem free from the influence of parameterization.

\subsection{Other models}
The Black-Scholes (BS) and Black-Scholes-Merton (BSM) models are world-renowned theoretical frameworks for option pricing, 
originally proposed by Black, Scholes, and Merton\cite{black1973pricing}\cite{merton1973theory}. 
These models serve as foundational cornerstones in the field of option pricing research.
However, within the domain of option pricing with machine learning, the BSM model has seen limited application—possibly due to its reliance on dividend yield data, 
which can be difficult to obtain. This changed with the work of Li et al., who incorporated the BSM model into a machine learning–based study\cite{li2024option}.

The Multilayer Perceptron (MLP) is a classical neural network model developed following the introduction of the backpropagation (BP) algorithm. 
It forms the fundamental logic of modern neural network architectures. 
Pinkus et al. demonstrated that neural networks—such as the MLP—can approximate continuous functions under suitable conditions, 
laying the theoretical foundation for their application\cite{pinkus1999approximation}.

Long Short-Term Memory (LSTM) networks are particularly effective for sequence prediction tasks, especially in time series forecasting. 
As a result, many researchers have employed LSTM models in machine learning–based option pricing studies\cite{ge20213d}\cite{ivașcu2021option}\cite{li2024option}.

Genetic Algorithms (GA) offer an alternative to traditional neural network architectures. 
They are designed to address complex problems that are not easily solved using conventional models. 
Ivașcu et al. found GA to be more suitable for option pricing tasks than many traditional approaches in their study\cite{ivașcu2021option}.

\section{Method}
\subsection{An experimental strategy}
Tuning is a common operation in machine learning that enables researchers to adjust hyperparameters to optimize model performance.
Consequently, it is widely adopted by scholars during experiments, as this strategy can effectively minimize errors.

From this perspective, employing a global tuning strategy appears to be a well-advised choice; 
however, given that option pricing falls within the domain of finance, the comprehensive ability of the models remains the most important consideration.

This paper proposes a novel hybrid tuning strategy, where hyperparameter tuning is conducted only during the initial sub-experiment within a major experiment. 
For some major experiments, the resulting parameter settings from the internal small experiment are adopted; for others, the parameter settings 
are directly inherited from preceding major experiments.
To further mitigate the risk of underestimating tuning effects, the selected hyperparameters are carefully determined.

For example, in the set of experiments presented in Table 2, hyperparameter tuning is conducted only in the experiment labeled In1, 
while the parameter settings of the remaining experiments in Table 6 are directly inherited from In1.

On the one hand, this approach improves consistency in parameter settings across experiments, while the input data in each experimental case remains distinct. 
This design enables a more reliable assessment of model robustness under varying data conditions. On the other hand, 
it aligns with the behavioral logic of financial practitioners—when an existing model structure proves effective, there is often no need to further optimize it. 
Similarly, if a particular parameter setting is appropriate for an experiment such as In1, it is reasonable to extend it to other related experiments.
\begin{table}[ht!]
  \begin{center}    
    \begin{tabular}{ccc} 
      \hline
      \textbf{Experiment} & \textbf{Corresponding Table} & \textbf{Parameter Setting} \\
      \hline
      Input Experiment & Table 2 & In1 \\
      Moneyness Experiment & Table 4 & ALL \\
      Sliding window Experiment & Table 6 & In1 \\
      Noise Experiment & Table 7 & ALL \\
      \hline
    \end{tabular}
    \caption{The experimental strategy used in this study}
  \end{center}
\end{table}
\subsection{An evaluation mechanism}
In the field of computer science, scholars place strong emphasis on model baselines, which serve as reference points for evaluating performance 
and facilitating comparisons between models.
Boxplots and performance curves are widely adopted by researchers in the area of option pricing using machine learning\cite{ge20213d}\cite{ivașcu2021option}\cite{li2024option}, 
as they align well with technical requirements and precision standards.
However, these forms of baseline expression primarily reflect the conventions of computer science and lack the theoretical flavor 
and domain relevance characteristic of financial research.

In this paper, a novel scoring strategy and a weighted evaluation strategy jointly form an evaluation mechanism that integrates elements of financial theory with the technical rigor of computer science.
This mechanism serves as the evaluation baseline for two of the four experiments conducted in the study. 
While all experiments include error metrics for performance assessment, 
the baseline mechanism is applied only where its comparative perspective is appropriate and meaningful.

The scoring strategy is defined by the following formulas:
\begin{equation}
\text{Score Rate} = \frac{E_{BS} - e}{E_{BS}} \times 100\%
\end{equation}

\begin{equation}
\text{Score Rate} = \frac{E - e}{E} \times 100\%
\end{equation}
where $E$ denotes the largest numerical error among the models within a sub-experiment (e.g., In3 in the input experiment), and
$E_{BS}$ represents the numerical error of the BS model in the same sub-experiment.
The variable $e$ refers to the smallest numerical error observed within that sub-experiment.

According to the definition, the two Score Rates are typically non-negative, 
unless the numerical error of the evaluated model is greater than that of the BS model. 
It is important to emphasize that a higher Score Rate always indicates better model performance.
Moreover, when comparing the Score Rates derived from the two formulas, a lower overall Score Rate in the presence of the BS model (i.e., using $E_{BS}$
) suggests that the performance gap among models has been reduced. This implies that the inclusion of the BS model contributes positively to the overall experiment.

Formulas (4) and (5) represent two versions of the proposed scoring strategy. Specifically, 
Formula (4) compares performance with the Black--Scholes (BS) model, while Formula (5) compares it with a machine learning model.
In the experimental section, the proposed formulas are applied in two major experiments to assess the performance of the models.

In the experimental section, the proposed formulas are applied in two major experiments to assess the performance of the models.

This scoring strategy applies the idea of model comparison broadly, extending it to incorporate the Black--Scholes (BS) model.
Compared to boxplots and performance curves, this strategy highlights the central role of the BS model and the comparative performance among models. 
Although boxplots impartially illustrate the relationship between errors and variance, they do not explicitly represent the financial models as core components.

The weighted evaluation strategy is developed based on the previously defined scoring strategy. Unlike ordinary equal-weighting schemes, 
the weights in this approach are assigned according to preferences derived from financial theory. 
For example, sub-experiments In3 and In4 in the input experiment are given higher weights because they correspond to inputs of the Black--Scholes (BS) model.

Specifically, in the input experiment, the weights assigned to the six sub-experiments are 1, 1, 2, 2, 1, and 1, respectively.
While using equal weights is a reasonable choice in general, such an approach fails to reflect financial priorities and reduces the distinctiveness of the evaluation mechanism in this case.
In contrast, the moneyness experiment assigns equal weights of 1 to each of its four sub-experiments, which is appropriate since this experiment does not involve financial-theoretical weighting considerations.
Together, the scoring strategy and the weighted evaluation strategy constitute the overall evaluation mechanism, the details of which are discussed in the Discussion section.

\subsection{Sliding window technique}
The sliding window technique is commonly employed in the field of time-series forecasting, which includes applications such as option pricing via machine learning. 
This method has been utilized in prior studies, including those by Ge et al.\cite{ge20213d} and Li et al.\cite{li2024option}.
Sliding windows are advantageous for time-series analysis as they can capture local information and generate corresponding local features. 
Furthermore, they offer flexibility in handling diverse datasets, allowing adaptation to specific task objectives.

In this paper, the sliding window is enabled in all experiments except the sliding window experiments, 
which are specifically designed to compare the performance with and without this technique.
\section{Experiment}
\subsection{Data and data processing method}
In the experiment, the CSI 300 Index option data collected from the Chinese financial markets are divided into two parts: the first spans from January 1st, 2020, to December 31st, 2020, and the second covers the period from September 1st, 2021, to December 31st, 2021.

The training dataset comprises data from January to August 2020, together with the newly added data from September to December 2021, while the test dataset consists of data from September to December 2020.
Incorporating the 2021 data into the training set is intended to enrich the training data with denoised information, enabling the construction of a noise-controlled training set aimed at evaluating the models’ performance in the presence of noise within the experimental framework.

In the data processing stage, this study adopts a strategy proposed by Li and Huang\cite{li2024option}, which aims to more accurately simulate real-world financial environments.

A total of eight variables are considered in this study, including strike price(K), spot price(S), time to maturity($\tau$), option type, risk-free rate(r), monthly dividend rate(q), Delta($\Delta$), and volatility calculated using the GARCH model($\sigma$).
These variables are either selected from or fully utilized, depending on the specific modeling configuration.

The training dataset makes use of up to 59,191 observations, while the test dataset utilizes up to 21,263 observations.
All data used in this analysis are obtained from www.resset.com.
\subsection{Experiments and results}
Four major experiments are conducted in this study to investigate the effects of input, moneyness, the sliding window, and noise. 
By employing the experimental strategy and the evaluation mechanism described in the Method section, 
these experiments demonstrate strong generalization capabilities and comprehensive performance characteristics.
\begin{table}[ht!]
  \begin{center}    
    \begin{tabular}{c c c c c c c} 
      \hline
      \textbf{   } & \textbf{In1} & \textbf{In2}& \textbf{In3} & \textbf{In4} & \textbf{In5} & \textbf{In6} \\
      \hline
      Input & {\footnotesize S, K, $\tau$, r} & {\footnotesize S/K, $\tau$, r} & {\footnotesize S, K, $\tau$, r, $\sigma$} & {\footnotesize S/K, $\tau$, r, $\sigma$} & {\footnotesize S, K, $\tau$, r, $\sigma$, $\Delta$} & {\footnotesize S/K, $\tau$, r, $\sigma$, $\Delta$} \\
      Output & C & C & C & C & C & C\\
      BS &0.1270\phantom{\textsuperscript{*}} &0.1270\phantom{\textsuperscript{*}}  &0.1270\phantom{\textsuperscript{*}} &0.1270\phantom{\textsuperscript{*}} &0.1270\phantom{\textsuperscript{*}} &0.1270\phantom{\textsuperscript{*}}\\
      BSM &0.1246\phantom{\textsuperscript{*}} &0.1246\phantom{\textsuperscript{*}}  &0.1246\phantom{\textsuperscript{*}} &0.1246\phantom{\textsuperscript{*}} &0.1246\phantom{\textsuperscript{*}} &0.1246\phantom{\textsuperscript{*}}\\
      CatBoost & 0.0988\phantom{\textsuperscript{*}} & 0.1427\phantom{\textsuperscript{*}} & 0.1295\phantom{\textsuperscript{*}} & 0.1569\phantom{\textsuperscript{*}} & 0.0858\phantom{\textsuperscript{*}} & 0.1491\phantom{\textsuperscript{*}}\\
      DeepForest & 0.1221\phantom{\textsuperscript{*}} & 0.0839\phantom{\textsuperscript{*}} & 0.1656\phantom{\textsuperscript{*}} & 0.0946\phantom{\textsuperscript{*}} & 0.1148\phantom{\textsuperscript{*}} & 0.0954\phantom{\textsuperscript{*}}\\
      GA & 0.1123\phantom{\textsuperscript{*}} & 0.0812\phantom{\textsuperscript{*}} & 0.1126\phantom{\textsuperscript{*}} & 0.0975\phantom{\textsuperscript{*}} & 0.0877\phantom{\textsuperscript{*}} & 0.0647\phantom{\textsuperscript{*}}\\
      LGBM &\textbf{0.0798}\textsuperscript{*} & 0.0801\phantom{\textsuperscript{*}} & 0.1125\phantom{\textsuperscript{*}} & 0.0744\phantom{\textsuperscript{*}} & \textbf{0.0620}\textsuperscript{*} & 0.0663\phantom{\textsuperscript{*}}\\
      LSTM & 0.0905\phantom{\textsuperscript{*}} & 0.1121\phantom{\textsuperscript{*}} & \textbf{0.0827}\textsuperscript{*} & 0.0924\phantom{\textsuperscript{*}} & 0.0858\phantom{\textsuperscript{*}} & 0.0931\phantom{\textsuperscript{*}}\\
      MLP & 0.1006\phantom{\textsuperscript{*}} & 0.0877\phantom{\textsuperscript{*}} & 0.1172\phantom{\textsuperscript{*}} & 0.0952\phantom{\textsuperscript{*}} & 0.0984\phantom{\textsuperscript{*}} & 0.0981\phantom{\textsuperscript{*}}\\
      NGBoost & 0.0840\phantom{\textsuperscript{*}} & 0.0820\phantom{\textsuperscript{*}} & 0.1182\phantom{\textsuperscript{*}} & 0.0806\phantom{\textsuperscript{*}} & 0.0655\phantom{\textsuperscript{*}} & 0.0763\phantom{\textsuperscript{*}}\\
      RF & 0.1126\phantom{\textsuperscript{*}} & 0.0812\phantom{\textsuperscript{*}} & 0.1123\phantom{\textsuperscript{*}} & 0.1025\phantom{\textsuperscript{*}} & 0.0864\phantom{\textsuperscript{*}} & \textbf{0.0645}\textsuperscript{*}\\
      XGBoost & 0.0859\phantom{\textsuperscript{*}} & \textbf{0.0770}\textsuperscript{*} & 0.1039\phantom{\textsuperscript{*}} & \textbf{0.0719}\textsuperscript{*} & 0.0686\phantom{\textsuperscript{*}} & 0.0651\phantom{\textsuperscript{*}}\\
      \hline
    \end{tabular}
    \caption{Error(RMSE) results of the input experiments.}
  \end{center}
\end{table}

\begin{table}[ht!]
  \begin{center}    
    \begin{tabular}{ccc} 
      \hline
      \textbf{Model} & \textbf{Score Rate with BS model} & \textbf{Score Rate with non-parameter model}\\
      \hline
      CatBoost & -3.2677 & 10.9929\\
      DeepForest & 7.8150 & 19.5794\\
      GA & 	24.5965 & 33.8829\\
      LGBM & 34.8425 & 43.6664\\
      LSTM & 27.9823 & 36.5603\\
      MLP & 20.3150 & 30.2181\\
      NGBoost & 30.5709 & 40.0021\\
      RF & 23.7894 & 33.2591\\
      XGBoost & 36.2008 & 44.3921\\
      \hline
    \end{tabular}
    \caption{Score Rate results of the input experiments.}
  \end{center}
\end{table}

Input experiments and moneyness experiments are among the most fundamental experiment types in the field of option pricing using machine learning.
Codruț-Florin et al.\cite{ivașcu2021option} introduced both types as essential components of their comprehensive study.
Moneyness is not only a key variable in the Black-Scholes-Merton (BSM) formula, but is also frequently emphasized in related research\cite{ivașcu2021option}\cite{ruf2019neural} 
for its critical role in the feild. In financial markets, investors commonly classify options based on their level of moneyness.
Studying moneyness in machine learning models represents an attempt to better align the research with real-world financial practices.

In this paper, moneyness is also highlighted within the design of the input experiments.
Specifically, input models In1 and In2 are designed as a comparable pair, differing in whether the spot price (S) and strike price (K) are provided as separate inputs, 
or combined into the moneyness ratio (S/K).
Other input models can similarly be grouped into comparable combinations. Each such combination serves as a basis for comparative analysis.

In the moneyness experiment, the dataset is divided into four subsets based on moneyness. 
\begin{table}[ht!]
  \begin{center}    
    \begin{tabular}{c c c c c c c} 
      \hline
      \textbf{   } & \textbf{ALL} & \textbf{ITM}& \textbf{ATM} & \textbf{OTM}\\
      \hline
      BS &0.1270\phantom{\textsuperscript{*}} &0.1423\phantom{\textsuperscript{*}}  &0.1954\phantom{\textsuperscript{*}} &0.0777\phantom{\textsuperscript{*}} \\
      BSM &0.1246\phantom{\textsuperscript{*}} &0.1176\phantom{\textsuperscript{*}}  &0.1910\phantom{\textsuperscript{*}} &0.0882\phantom{\textsuperscript{*}}\\
      CatBoost & 0.0799\phantom{\textsuperscript{*}} & 0.0838\phantom{\textsuperscript{*}} & 0.0467\phantom{\textsuperscript{*}} & 0.1904\phantom{\textsuperscript{*}}\\
      DeepForest & 0.0878\phantom{\textsuperscript{*}} & 0.0611\phantom{\textsuperscript{*}} & 0.0400\phantom{\textsuperscript{*}} & 0.1031\phantom{\textsuperscript{*}}\\
      GA & 0.0714\phantom{\textsuperscript{*}} & 0.0436\phantom{\textsuperscript{*}} & 0.0419\phantom{\textsuperscript{*}} & 0.0454\phantom{\textsuperscript{*}}\\
      LGBM &\textbf{0.0385}\textsuperscript{*} & 0.0406\phantom{\textsuperscript{*}} & \textbf{0.0346}\textsuperscript{*} & \textbf{0.0428}\textsuperscript{*}\\
      LSTM & 0.1264\phantom{\textsuperscript{*}} & 0.1569\phantom{\textsuperscript{*}} & 0.0817\phantom{\textsuperscript{*}} & 0.1252\phantom{\textsuperscript{*}}\\
      MLP & 0.0910\phantom{\textsuperscript{*}} & 0.0863\phantom{\textsuperscript{*}} & 0.0650\phantom{\textsuperscript{*}} & 0.1151\phantom{\textsuperscript{*}}\\
      NGBoost & 0.0629\phantom{\textsuperscript{*}} & 0.0388\phantom{\textsuperscript{*}} & 0.0347\phantom{\textsuperscript{*}} & 0.0485\phantom{\textsuperscript{*}}\\
      RF & 0.0670\phantom{\textsuperscript{*}} & 0.0423\phantom{\textsuperscript{*}} & 0.0417\phantom{\textsuperscript{*}} & 0.0453\phantom{\textsuperscript{*}}\\
      XGBoost & 0.0438\phantom{\textsuperscript{*}} & \textbf{0.0376}\textsuperscript{*} & 0.0379\phantom{\textsuperscript{*}} & 0.0434\phantom{\textsuperscript{*}}\\
      \hline
    \end{tabular}
    \caption{Error(RMSE) results of the moneyness experiments.}
  \end{center}
\end{table}

\begin{table}[ht!]
  \begin{center}    
    \begin{tabular}{ccc} 
      \hline
      \textbf{Model} & \textbf{Score Rate with BS model} & \textbf{Score Rate with non-parameter model}\\
      \hline
      CatBoost & 2.3131 & 31.5545\\
      DeepForest & 33.6920 & 47.1218\\
      GA & 58.3167 & 60.1486\\
      LGBM & 67.0907 & 69.7089\\
      LSTM & -3.1830 & 8.5609\\
      MLP &	21.5752 & 33.2480\\
      NGBoost & 60.7570 & 64.3908\\
      RF & 59.4690 & 61.3004\\
      XGBoost & 65.9592 & 68.0501\\
      \hline
    \end{tabular}
    \caption{Score Rate results of the moneyness experiments.}
  \end{center}
\end{table}

ALL refers to the entire dataset. 
ATM (At-the-Money) corresponds to options with a moneyness ratio (S/K) in the interval $[0.96, 1.04]$. 
ITM (In-the-Money) includes options where S/K $\in (0, 0.96)$, and 
OTM (Out-of-the-Money) includes those where S/K $\in (1.04, \infty)$.
The variables selected for the other three major experiments are consistent: S/K, $\tau$, r, $\sigma$, and q. 
The variable q was first introduced by Li and Huang in their study on option pricing using machine learning\cite{li2024option}.
Details of these variables are provided in the Data and data processing method subsection.

The results of the input experiments and moneyness experiments are shown in Table 2 and Table 4.
\begin{table}[ht!]
  \begin{center}    
    \begin{tabular}{c|cc|cc|cc} 
      \hline
      \textbf{} & \multicolumn{2}{c|}{\textbf{ITM}} & \multicolumn{2}{c|}{\textbf{ATM}} & \multicolumn{2}{c}{\textbf{OTM}} \\
      \textbf{} & ON & OFF & ON & OFF & ON & OFF \\
      \hline
      CatBoost & 0.00702 & \textbf{0.00461} & 0.00203 & \textbf{0.00188} & 0.02717 & \textbf{0.02689} \\
      DeepForest &  \textbf{0.00240} & 0.00367 & 0.00165 & \textbf{0.00123} & \textbf{0.00704} & 0.01341 \\
      GA & 0.00180 & \textbf{0.00123} & \textbf{0.00166} & 0.00217 & \textbf{0.00208} & 0.00234 \\
      LGBM & 0.00165 & \textbf{0.00143} & 0.00120 & \textbf{0.00105} & \textbf{0.00183} & 0.00251 \\
      LSTM & 0.02806 & \textbf{0.01744} & 0.00884 & \textbf{0.00777} & \textbf{0.01311} & 0.01629 \\
      MLP & 0.03245 & \textbf{0.01480} & 0.00828 & \textbf{0.00520} & 0.01542 & \textbf{0.01415} \\
      NGBoost & \textbf{0.00150} & 0.00187 & 0.00121 & \textbf{0.00111} & 0.00235 & \textbf{0.00220} \\
      RF & 0.00186 & \textbf{0.00124} & \textbf{0.00174} & 0.00211 & \textbf{0.00209} & 0.00226 \\
      XGBoost & 0.00141 & \textbf{0.00109} & 0.00143 & \textbf{0.00139} & \textbf{0.00189} & 0.00235 \\
      \hline
    \end{tabular}
    \caption{Error(RMSE) results of the sliding window experiments.}
  \end{center}
\end{table}

\begin{table}[ht!]
  \begin{center}    
    \begin{tabular}{cccc} 
      \hline
      \textbf{Model} & \textbf{Original Dataset} & \textbf{Denoised Dataset} & \textbf{Denoised Error Increase (\%)} \\
      \hline
      CatBoost & 0.01645 & 0.03934 & 139.15 \\
      DeepForest & 0.02691 & 0.05495 & 104.20 \\
      GA & 0.01275 & 0.02856 & 124.00 \\
      LGBM & 0.01207 & 0.01284 & 6.38 \\
      LSTM & 0.00933 & 0.02982 & 219.61 \\
      MLP & 0.02457 & 0.00821 & -66.59 \\
      NGBoost & 0.00826 & 0.04214 & 410.17 \\
      RF & 0.01263 & 0.03314 & 162.39 \\
      XGBoost & 0.00987 & 0.01470 & 48.94 \\
      \hline
    \end{tabular}
    \caption{Error(MSE) results of the noise experiments.}
  \end{center}
\end{table}

The sliding window experiments and noise experiments are designed as comparative studies, but they can also be interpreted and analyzed independently.
The results are expressed in Table 6 and Table 7.

The sliding window technique is effective in capturing local feature information and possesses several advantageous properties, as described in the Methods section.
The objectives of the sliding window experiments are twofold: to evaluate the models’ capacity to receive and process local feature information,
and to assess their performance under varying moneyness conditions.

Noise is inherent in most datasets and can hinder experts from obtaining accurate results.
Therefore, it is essential to design dedicated experiments to explore the impact of noise.
The noise experiments, based on the dataset constructed in the Data and data processing method section, 
are designed to evaluate the models’ ability to counteract the effects of noise.

All the experiments are discussed carefully in the Discussion section.
\section{Discussion}
In this section, the content of the Method and Experiment sections is reviewed and discussed to help readers better understand the details.

In the Method section, the experimental strategy and evaluation mechanism are introduced for the first time. 
Table 1 presents the specific parameter settings that reflect the experimental strategy employed in this study.

For the input experiment and the moneyness experiment, the parameter tuning results of In1 and ALL are applied to their respective major experiments, 
which is a reasonable choice. However, it raises the question: why are In1 and ALL also selected as the parameter settings for the sliding window experiment and the noise experiment, respectively?
In other words, since the input of the sliding window experiment is closely related to moneyness, would it not be more appropriate to select ALL instead of In1?

The answer to this question lies in the core objective of this paper: to explore the application of financial theory and the generalization capability of models, 
rather than focusing solely on accuracy or advanced computational techniques.
While both directions are valid, it is often difficult—if not impossible—for researchers to simultaneously prioritize both accuracy and generalization in a single study.

For this reason, we adopt the parameter settings presented in Table 1.
With regard to the evaluation mechanism, both the scoring strategy and the weighted strategy are closely aligned with financial theory and emphasize 
the idea of combination, reflecting an exploratory mindset in experimental design.

In the Experiment section, we conduct four major experiments, which will be discussed in order.
The input experiment is carefully designed with comparable input combinations, aiming to investigate the adaptability of models to moneyness-related aspects.
Table 2 indicates that most models show better performance when moneyness (S/K) is included as an input, compared to using S and K separately.
The moneyness experiment follows standard conventions and is divided into four sub-experiments based on different datasets.
Differ from input experiment, all sub-experiments share the same set of input parameters.

The applications of the evaluation mechanism are presented in Table 3 and Table 5.
As emphasized in the corresponding subsection, a higher Score Rate consistently indicates better model performance.

The necessity of incorporating the BS model is clearly demonstrated, as the Score Rate is significantly lower when it is included compared to when it is not.
This implies that the BS model contributes positively to the experiment—if it were ineffective, its inclusion would not result in a lower Score Rate.

In terms of capability and performance among non-parametric models, the ensemble methods—LGBM, NGBoost, and XGBoost—consistently outperform other models in both major experiments.
The GA model also demonstrates superior performance in comparison to the traditional MLP model.
Although CatBoost performs poorly in the input experiment, it yields better results in the moneyness experiment.
DeepForest also demonstrates good stability across both experiments.

The debate surrounding the flexibility and noise robustness of models is well-known, yet remains uncertain and inconsistent across studies.
The sliding window experiment and the noise experiment serve as representative tests of flexibility and noise resilience, respectively, 
as sliding window techniques are often employed to capture local information in model design\cite{gou2020sliding}.
These two major experiments are constructed in a comparative manner to explore the deeper relationship between flexibility and noise robustness in modeling.

First, focusing on the two experiments separately:
Table 6 indicates that the sliding window setting is not effective under ITM and ATM conditions, whereas it becomes a reasonable choice in the OTM condition.
This suggests that the applicability of the sliding window technique may vary depending on the moneyness type.
Table 7 presents the noise robustness of the models, with the MLP model demonstrating exceptional performance in this regard.
This can be attributed to its specific architecture, which enables better performance when the amount of input data increases, 
provided that the increase remains within a suitable range.
The design of a denoised dataset aligns well with this characteristic of the MLP model. When it comes to other models,
both LGBM and XGBoost exhibit strong and consistent performance, whereas NGBoost demonstrates considerable fluctuations across experiments.

Then based on the results of these experiments, it is not appropriate to conclude that model flexibility inherently compromises noise robustness, or vice versa.
The idea that flexibility and noise robustness are inherently opposed remains more of a research aspiration than an established fact.
In fact, the experimental results suggest a potential trade-off between these two properties. For instance, 
LGBM and XGBoost perform better without sliding windows in ITM and ATM conditions, yet show strong performance in the noise experiment, 
indicating a possible tension between flexibility and noise robustness.
However, this observation is not sufficient to definitively confirm or refute such a relationship. As such, 
it remains premature to conclude whether flexibility and noise robustness are inherently opposed or compatible.

\section{Conclusion}

In this study, four carefully designed major experiments are conducted to evaluate models’ performance across multiple dimensions, including precision, 
flexibility, noise robustness, and compatibility with financial theory. Ensemble learning models such as LGBM, XGBoost, 
and NGBoost demonstrate outstanding performance in these exploratory settings. 
These results are derived from a novel experimental strategy that emphasizes generalization capability over purely accuracy-driven approaches.

A new evaluation mechanism is proposed, in which the score rates observed in the input experiment and moneyness experiment highlight 
the importance of incorporating classical financial theory—specifically, the Black-Scholes (BS) model—as its inclusion effectively reduces the score rate, 
indicating its positive influence. This evaluation strategy not only provides a novel baseline for model evaluation in computational finance 
but also invites future refinement and adaptation by other experts in the field.

The sliding window experiment and the noise experiment are designed to be both comparative and independently analyzable. 
Whether flexibility and noise robustness are inherently contradictory characteristics remains an open question worthy of continued investigation. 
While certain results suggest a potential trade-off between the two, others fail to provide definitive evidence. 
This topic invites further exploration through carefully designed experiments by future scholars.

The results of the sliding window experiment indicate that model performance with sliding windows—interpreted as a measure of flexibility—varies depending 
on the type of moneyness. Meanwhile, the findings from the noise experiment shed light on the models’ ability to cope with noisy data, 
offering meaningful benchmarks for comparison. Together, these results provide valuable insights and guidance for future research in the domain of option pricing.
\section{Acknowledgments}
This research was supported by National Natural Science Foundation of China(12171196).





\bibliographystyle{elsarticle-harv}
\bibliography{ref}

\end{document}